\documentclass{article}

\usepackage{arxiv}

\usepackage[utf8]{inputenc} 
\usepackage[T1]{fontenc}    
\usepackage{hyperref}       
\usepackage{url}            
\usepackage{booktabs}       
\usepackage{amsfonts}       
\usepackage{nicefrac}       
\usepackage{microtype}      
\usepackage{lipsum}         
\usepackage{graphicx}       
\usepackage{natbib}         
\usepackage{doi}            
\usepackage{authblk}        
\usepackage{amsmath}        
\usepackage{array}          
\usepackage[table,xcdraw,dvipsnames]{xcolor} 
\usepackage{soul}           
\usepackage{geometry}       
\usepackage{tabularx}       
\usepackage{multirow}       
\usepackage{longtable}      

\title{SEMIKONG: CURATING, TRAINING, AND EVALUATING \\ A SEMICONDUCTOR INDUSTRY-SPECIFIC \\ LARGE LANGUAGE MODEL}

\date{} 					

\author{\large Christopher Nguyen\textsuperscript{1}, William Nguyen\textsuperscript{1}, Atsushi Suzuki\textsuperscript{3}, Daisuke Oku\textsuperscript{3}, Hong An Phan \textsuperscript{1},\\ Sang Dinh\textsuperscript{1}, Zooey Nguyen\textsuperscript{1}, Anh Ha\textsuperscript{1},
    \large Shruti Raghavan\textsuperscript{1}, Huy Vo\textsuperscript{2}, Thang Nguyen\textsuperscript{2},
    \large Lan Nguyen\textsuperscript{2}, Yoshikuni Hirayama}

\affil[1]{Aitomatic, Inc.}
\affil[2]{FPT Software, AI Center}
\affil[3]{Tokyo Electron Ltd}

\renewcommand{\shorttitle}{SemiKong: Curating, Training, and Evaluating A Semiconductor Industry-specific LLM}

\hypersetup{
pdftitle={A template for the arxiv style},
pdfsubject={q-bio.NC, q-bio.QM},
pdfauthor={David S.~Hippocampus, Elias D.~Striatum},
pdfkeywords={First keyword, Second keyword, More},
}

\begin{document}
\maketitle

\begin{abstract}
    Large Language Models (LLMs) have demonstrated the potential to address some issues within the semiconductor industry. However, they are often general-purpose models that lack the specialized knowledge needed to tackle the unique challenges of this sector, such as the intricate physics and chemistry of semiconductor devices and processes. SemiKong, the first industry-specific LLM for the semiconductor domain, provides a foundation that can be used to develop tailored proprietary models. With SemiKong 1.0, we aim to develop a foundational model capable of understanding etching problems at an expert level. Our key contributions include (a) curating a comprehensive corpus of semiconductor-related texts, (b) creating a foundational model with in-depth semiconductor knowledge, and (c) introducing a framework for integrating expert knowledge, thereby advancing the evaluation process of domain-specific AI models. Through fine-tuning a pre-trained LLM using our curated dataset, we have shown that SemiKong outperforms larger, general-purpose LLMs in various semiconductor manufacturing and design tasks. Our extensive experiments underscore the importance of developing domain-specific LLMs as a foundation for company- or tool-specific proprietary models, paving the way for further research and applications in the semiconductor domain. Code and dataset will be available at \url{https://github.com/aitomatic/semikong}\footnote{\url{https://www.semikong.ai}}.

\end{abstract}


\keywords{Semiconductor industry \and Domain-specific language model \and Industry-specific corpus \and Adaptive pre-training \and Manufacturing and design tasks \and Process optimization \and Defect detection \and Open models}

\section{Introduction}

\subsection{Semiconductor Manufacturing and Design}

Semiconductors play an essential role in powering various electronic devices and driving development across industries such as telecommunications, automotive, healthcare, renewable energy, and IoT. In semiconductor manufacturing and design, the two main phases, FEOL and BEOL, each present their own unique challenges. FEOL, the front end of line processes, involves the creation of active devices on the semiconductor wafer. This includes steps such as wafer preparation, photolithography, etching, ion implantation, and gate oxide formation~\cite{ElKareh1994FundamentalsOS}. These processes are crucial for defining the transistor structures and other active components of the integrated circuit (IC)~\cite{Xiao2000IntroductionTS}. On the other hand, BEOL, the back end of line processes, focuses on connecting the active devices created during FEOL. This includes the formation of metal layers, insulation, and bonding pads~\cite{quirk2001semiconductor}. Back-end processes are essential for establishing the electrical connections between devices and enabling the overall functionality of the IC~\cite{May2006FundamentalsOS}.
As feature sizes continue to shrink and device architectures become more complex, the need for advanced manufacturing techniques and design methodologies has become paramount. This has led to a growing interest in leveraging artificial intelligence (AI) and machine learning (ML) techniques to optimize semiconductor manufacturing processes and assist in design tasks~\cite{Amuru2022AIMLAA, 2022MachineLA, Huang2021MachineLF}.

\subsection{Use of LLMs in Semiconductors}

Recent advancements in LLMs have demonstrated their remarkable potential in various domains, including the semiconductor industry ~\cite{Liu2023ChipNeMoDL, Liu2023RTLCoderOG}. LLMs, trained on vast amounts of text data using self-supervised learning techniques, have shown the ability to capture rich domain knowledge and generate human-like text. This has opened up new possibilities for applying LLMs to semiconductor process technology and IC design tasks.
In the context of semiconductor process technology, LLMs can potentially assist in tasks such as process parameter optimization~\cite{Liu2024LargeLM}, anomaly detection~\cite{RussellGilbert2024AADLLMAA}, and predictive maintenance of manufacturing equipment~\cite{Lee2023AUI}. By leveraging the vast amount of process data and domain knowledge embedded in the pre-trained models, LLMs can help identify patterns, predict process outcomes, and suggest optimal settings for various manufacturing steps.
Similarly, in the realm of IC design, LLMs can aid in tasks such as design rule checking, layout generation, and design space exploration~\cite{Chang2023ChipGPTHF}. By learning from large datasets of IC layouts and design rules~\cite{Mallappa2024FloorSetA}, LLMs can potentially generate new designs that adhere to the specified constraints and optimize for desired performance metrics.

\begin{figure}[!htbp]
\centering
\includegraphics[width=1.0\textwidth]{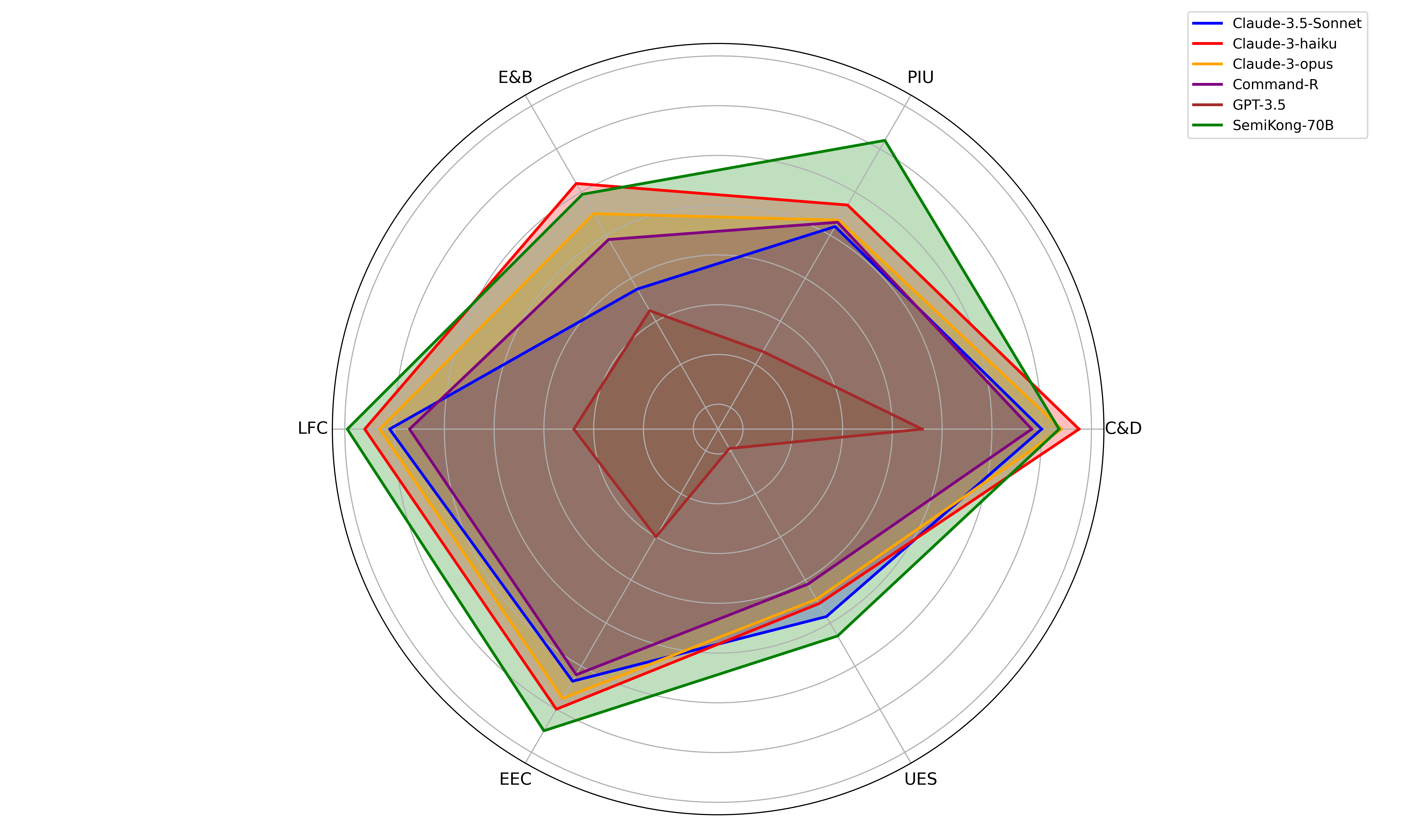} 
\caption{\textbf{Comparison of SemiKong and commercial models.} SemiKong is a open source foundation model but achieved comparable performance on E\&B (Efficiency and Brevity), C\&D (Clarity and Directness) with other commercial models and significantly outperformed these products in PIU (Practicality and Immediate Usability), LFC (Logical Flow and Coherence), EEC (Expert-to-Expert Communication), UES (Use of Examples and Specificity).}
\label{fig:products_comparison}
\end{figure}

\subsection{Purpose and Scope}

Building on the success and potential of LLMs, this paper introduces SemiKong, the first industry-specific LLM tailored for the semiconductor domain, focusing on applications in semiconductor process technology and manufacturing. We aim to address the limitations of generic foundation models by curating a comprehensive semiconductor-related text corpus and developing a novel pre-training approach that leverages domain-specific knowledge. By doing so, we seek to demonstrate the potential of industry-specific LLMs in improving the performance of AI-driven solutions for semiconductor manufacturing tasks.

The scope of this work encompasses the following:
\begin{itemize}
 \item The curation of a large-scale, semiconductor-specific text corpus focused on process technology and manufacturing
 \item The development of SemiKong, a foundation model, specifically focuses on the etching problems in the semiconductor industry
 \item The fine-tuning of SemiKong on industry-relevant data and tasks related to process optimization and control
 \item The introduction of a novel framework to leverage expert feedback in order to advance the LLMs-based evaluation approach for domain-specific AI models.
 \item The evaluation of SemiKong's performance compared to general-purpose LLMs
 \item The discussion of the implications and potential applications of industry-specific LLMs in semiconductor manufacturing
\end{itemize}

The main contributions of this paper are as follows:
\begin{itemize}
    \item SemiKong-Corpus: We curate a comprehensive semiconductor-related text corpus, covering a wide range of topics related to semiconductor process technology and manufacturing. This corpus serves as the foundation for training SemiKong and captures the domain-specific knowledge essential for addressing manufacturing-related tasks.
    \item SemiKong-Trainer: We present SemiKong, a specialized foundation model with extensive knowledge of semiconductor manufacturing terminology and process flows, with a particular focus on etching. By pretraining and fine-tuning SemiKong with our carefully curated data, we have achieved substantial quality improvements in downstream tasks compared to generic LLMs and even commercial LLM-based products, as demonstrated in Figure~\ref{fig:products_comparison}.
    \item SemiKong-Eval: We develop a novel framework to effectively leverage expert’s knowledge to advance the LLMs-based evaluation process and produce high-quality benchmarks. Besides, we conduct extensive evaluations to assess SemiKong's performance on industry-relevant benchmarks, such as process parameter optimization, anomaly detection, and predictive maintenance. Our results demonstrate SemiKong's superiority over general-purpose LLMs, highlighting the importance of developing industry-specific models for the semiconductor manufacturing domain.
\end{itemize}

The remainder of this paper is organized as follows: Section~\ref{related_work} provides an overview of related work on the application of AI and LLM in the semiconductor industry. Section~\ref{ontology_section} introduces the semiconductor ontology, with a focus on the front-end processes of semiconductor manufacturing. Section~\ref{method_section} outlines the methodology used to curate a semiconductor-specific text corpus and develop the pre-training approach. Section~\ref{experiment_section} presents the experimental setup and results, comparing the performance of SemiKong with general-purpose LLMs across various manufacturing tasks. Section~\ref{discussion_section} discusses the implications of the findings, potential future research directions, and concludes the paper.

\section{Related Works}
\label{related_work}
\subsection{Artificial Intelligence (AI) in semiconductor manufacturing}

The application of Artificial Intelligence (AI) in semiconductor manufacturing has seen significant advancements, leveraging various AI methods to enhance the efficiency, yield, and quality of semiconductor fabrication processes. This section reviews the state-of-the-art AI approaches applied in different stages of semiconductor manufacturing, including two important steps: mask optimization, and hotspot detection.
Mask optimization is a critical step in semiconductor manufacturing. Traditional mask optimization methods typically consume significant runtime due to their iterative characteristics~\cite{Gu2008OpticalPC,Hung2002HybridOP,Kotani2002AdvancedHO}. Recently, machine learning-based methods are proposed to accelerate mask optimization tasks~\cite{Luo2013OpticalPC,Matsunawa2015OpticalPC,Choi2016MachineL}. ILILT~\cite{Yang2024ILILTIL} applied implicit learning for inverse lithography methods in mask optimization tasks. A large dataset, LithoBench~\cite{Zheng2023LithoBenchBA}, consists of more than 120k circuit layout tiles for deep learning-based lithography simulation and mask optimization and is published to accelerate machine learning-based approaches. In addition, in the task of mask optimization, deep reinforcement learning is proposed to be applied to directly optimize the preferred objective in optical proximity correction (OPC)~\cite{Liang2024RLOPCMO}. CAMO~\cite{Liang2024CAMOCM} , a modulated reinforcement learning for correlation-aware mask optimization, is proposed to exploit the spatial correlation between the movements of neighboring segments.

Hotspot detection is an important step in semiconductor manufacturing to ensure the reliability and performance of integrated circuits (ICs). Hotspots are areas on a chip where excessive heat or stress can lead to defects, reducing yield and affecting the longevity and functionality of the devices. With the continuous scaling down of semiconductor technology nodes, the detection and mitigation of these hotspots have become increasingly significant. An active learning-based hotspot detection method~\cite{Yang2018BridgingTG} achieved an impressive performance in terms of detection accuracy. A new lithography hotspot detection framework based on the AdaBoost classifier and a simplified feature extraction~\cite{Matsunawa2015ANL} obtained high accuracy with very low false alarms. In addition, semi-supervised learning with self-paced multi-task learning ~\cite{Chen2019SemiSupervisedHD} is proposed for hotspot detection. Meanwhile, a hotspot detection using deep convolutional neural networks~\cite{Shin2016AccurateLH} obtained accurate detection performance. These methods just focus on specific tasks rather than building a model to comprehensively support semiconductor operation engineers.

\subsection{LLMs in the semiconductor industry}

LLMs are proposed to adapt to domain-specific chip design, consisting of a wide range of tasks, from code generation to bug summarization and chatbot assistance for EDA engineers. ChipNemo~\cite{MLiu2023ChipNeMoDL} developed by NVIDIA, proved that domain fine-tuned LLM models outperform general-purpose LLM models such as Llama3, and GPT4 in three specific tasks as engineering assistant chatbot for Q\&A, EDA scripts generation, and bug summarization and analysis. RTLCoder~\cite{SLiu2023RTLCoderOG} outperforms GPT-3.5 in design RTL generation with an open-source dataset and a new training scheme via code quality feedback. ChipGPT~\cite{Chang2024DataIA} reinforces data-driven methods by making clear that data is all you need to finetune an LLM model for chip design, the results demonstrate a significant improvement in the code generation tasks with domain LLMs. Hdldebugger~\cite{Yao2024HDLdebuggerSH} focuses on using the LLM model for debugging via the LLM-assisted HDL debugging framework. Meanwhile, Rtlfixer~\cite{Tsai2023RTLFixerAF} targets fixing RTL syntax errors automatically with LLM models.  Chip-Chat~\cite{Blocklove2023ChipChatCA} conducted experiments with conversational LLMs to design and verify an 8-bit accumulator with GPT-4 and GPT-3.5. ChatEDA~\cite{He2023ChatEDAAL} introduces an autonomous agent for EDA empowered by a fine-tuned LLaMA2 70B model that outperforms the GPT-4 model in this task. In addition, inspired by LLMs in Natural Language Processing (NLP),  Large Circuit Models~\cite{Chen2024TheDO} are proposed as a new paradigm to streamline the EDA process. However, these models are mostly developed with small public datasets with the limitation of the expert's participation in the development process.

\subsection{LLM as an evaluator}

Human evaluation is a crucial method for assessing Natural Language Generation (NLG) algorithms~\cite{Guzmn2015HowDH, Gillick2010NonExpertEO}. Many NLP tasks require skilled annotators or experts for reliable evaluations~\cite{Gillick2010NonExpertEO}. However, recruiting human experts is often impractical due to high costs and concerns about reproducibility. Meanwhile, automatic metrics like BLEU~\cite{Papineni2002BleuAM} and ROUGE~\cite{Lin2004ROUGEAP} fall short of reliability expectations and fail to reflect human preferences accurately. Recently, using LLMs to evaluate NLG~\cite{Chiang2023CanLL, Fu2023GPTScoreEA, Wang2023IsCA} has been introduced to address these issues. These methods are reference-free, asking LLMs to justify their answers based on task requirements and demonstrating correlation with human judgment, assuming that LLMs can understand and assign higher probabilities to high-quality and fluent texts. G-eval~\cite{Liu2023GEvalNE} applied the chain-of-thought technique by asking LLMs to generate detailed evaluation steps to enhance evaluation quality. Despite these advancements, these methods share a common limitation: they assume LLMs can inherently understand and evaluate knowledge. However, experts with many years of experience are often needed to evaluate complex questions in domains requiring deep expertise, such as semiconductors, for accurate judgments.
Given these challenges, this paper proposes a framework that leverages expert feedback to create criteria for more reliable assessments by LLMs, approaching expert-level reliability. This feedback is also used to generate a high-quality benchmark for the semiconductor domain. OSCaR~\cite{Nguyen2024OSCaROS} employed a similar approach in generating high-quality benchmarks. However, they utilized feedback from normal humans on Amazon MTurk, while our benchmark relies on expert knowledge, ensuring significantly higher reliability.

\section{Semiconductor Ontology}
\label{ontology_section}

Semiconductor manufacturing involves numerous complex steps and processes, requiring extensive knowledge for effective execution. In each step, having an expert specialized in that particular field to guide workers is crucial. However, the semiconductor manufacturing process is not easily accessible to AI researchers, who possess deep expertise in AI but often lack domain-specific knowledge, particularly an understanding of semiconductor manufacturing. This gap hinders the development of efficient, domain-specific AI models. To address this challenge, we collaborated with semiconductor experts to develop an ontology that systematically structures the entire semiconductor manufacturing process. This ontology is constructed using a top-down approach, dividing the field from general to detailed levels, sub-levels, and specific processes, ensuring that no critical process is overlooked.

By systematically structuring the semiconductor manufacturing process, our ontology not only addresses the knowledge gap for AI researchers but also serves as a foundation for creating more effective domain-specific AI models. This ontology is invaluable not only for building specialized AI models, like SemiKong for etching, but also serves as a benchmark for evaluating future general intelligence models that aim to address a wide range of semiconductor manufacturing topics, both in model development and evaluation. The hierarchical structure of the ontology enhances understanding and training efficiency, enabling the creation of specialized language model agents with precise insights tailored to specific stages of semiconductor manufacturing. Consequently, this ontology serves as a dynamic tool for guiding future training efforts and ensuring that language models remain up-to-date with industry advancements. To achieve these objectives, a well-designed procedure and meticulous implementation are essential in constructing a comprehensive semiconductor ontology.

Our ontology for semiconductor manufacturing was developed in collaboration with industry experts to cover the entire semiconductor manufacturing process, from front-end to back-end, including Substrate Preparation, Film Formation, Patterning, Doping, Planarization, Cleaning and Surface Preparation, Thermal Processing, Metrology and Inspection, Advanced Modules, and Back-End Processes. These represent the primary levels of semiconductor fabrication, which our experts further divided into secondary and tertiary levels. For example, Patterning is a key first-level process, which is further broken down in the second level into subclasses such as Etching. The third level categorizes Etching into Wet Etching, Dry Etching, Plasma Etching, Reactive Ion Etching, Deep Reactive Ion Etching, Isotropic Wet Etching, Anisotropic Wet Etching, Atomic Ion Etching, and Electron Cyclotron Etching. This paper introduces our model, SemiKong, which can comprehensively understand and provide support for the etching process, ensuring that our ontology fully covers this critical area and lays the groundwork for future specialized models in other semiconductor manufacturing processes. Our ontology is detailed in section~\ref{ontology_section_appendix} in our appendix.

\section{SemiKong: Semiconductor Industry Specific LLM}
\label{method_section}

Developing an expert-level, domain-specific model necessitates acquiring in-depth knowledge in the relevant field. A prevalent approach involves training models with comprehensive domain-specific data. This training process can be divided into two stages: pretraining and fine-tuning. Although this method typically leads to significant model improvements, it still presents challenges related to data quality assurance, defining the model training strategy, and determining appropriate evaluation metrics. In this section, we will discuss our data curation pipeline (Section~\ref{data_curation_section}), the process for training the SemiKong model using both pretraining and fine-tuning (Section~\ref{model_training_section}), and the incorporation of expert feedback in the evaluation pipeline (Section~\ref{expert_feedback_section}).

\subsection{Data Curation}
\label{data_curation_section}

High-quality domain-specific datasets, including those for the semiconductor domain, are often scarce. To address this problem, we introduce a large-scale, high-quality text-based dataset specifically for the semiconductor domain. Our dataset consists of two parts: documents for pretraining and instructions for fine-tuning.

Pre-Training dataset: Pretraining is a crucial step for incorporating knowledge into models. However, pretrained generic models often prioritize data coverage over depth. It is challenging to determine which data was used to train the model and the extent of the knowledge it encompasses. Based on this issue, we assume that generic pretrained models lack in-depth knowledge and the ability to focus on specific domains. We introduce a text-based dataset focused on semiconductors, extracted from technical books, papers, and patents. To construct this dataset, we manually searched for public PDF documents available on the internet. These documents were then converted to raw text using the PyPDF library. Since the raw text often has formatting issues, we employed GPT-4o-mini for post-processing to transform the text into markdown format. This step not only corrected parsing errors but also preserved special types of information, such as tables. The effectiveness of our proposed pre-training dataset is demonstrated in the experimental results shown in Table IV. The results indicate significant improvement when comparing a model purely fine-tuned with instructions to a model pre-trained with our dataset before fine-tuning.

Instruction dataset: We utilized GPT-4o and GPT-o1-preview to generate instructions related to semiconductor keywords. To achieve this, we began by predefining a list of semiconductor-related terms, which guided GPT-4o in generating additional synonyms and related keywords. This expanded list was then used to direct GPT-4o in formulating questions for our dataset. Our approach ensures comprehensive coverage of issues that our SemiKong can address, thereby enhancing the effectiveness of our instruction dataset. The dataset comprises 5,000 questions explaining semiconductor concepts, 5,000 questions solving complex etching problems requiring mathematical reasoning, and 40,000 questions addressing standard etching process issues as shown in Table I. Once the question set was completed, we used GPT-4o to answer questions related to semiconductor concepts and routine problems. For more complex questions involving math and reasoning, we employed GPT-o1-preview to generate answers. This approach strengthens the model's capability in solving complex problems, making it a more robust foundation model, particularly in the semiconductor etching domain.

\begin{table}[h!]
\centering
\caption{Details of SemiKong dataset}
\begin{tabularx}{0.8\textwidth}{>{\raggedright\arraybackslash}X>{\centering\arraybackslash}X}
\toprule
\textbf{Dataset Details} & \textbf{Quantity} \\ \midrule
Number of books, book chapters & 129 \\ 
Number of etching-specific research papers & 708 \\ 
Number of research papers & 20K \\ 
Number of instructions & 50K \\  
Number of tokens & 525.6M \\ 
\bottomrule
\end{tabularx}
\label{table:datasets}
\end{table}

\subsection{Model tranining}
\label{model_training_section}

The curated dataset described in Section 3.1 was employed to train our SemiKong models. Initially, the text data was tokenized using Tiktoken, a tokenizer based on BPE, which is widely utilized in numerous NLP applications. Subsequently, Rotary Position Embedding (RoPE) was incorporated as the positional embedding component to enable the LLM to capture positional information effectively. The training process comprised two stages: model pre-training using our pure text dataset and supervised fine-tuning (SFT). Then, we do post-training processing to make models become more suitable for production. The model overview and computational resources are detailed in Table II.

\textbf{Model pre-training:} We hypothesized that generic pre-trained models lack domain-specific knowledge. Therefore, we pre-trained our SemiKong models using the Llama3 8B and 70B checkpoints from Meta as a starting point. This step aims to enhance the models' in-depth semiconductor domain-specific knowledge, thereby ensuring that they focus more on the specific domain in which we intend for the models to serve as experts in the future.

\textbf{Supervised fine-tuning (SFT):} While pre-training equips the models with in-depth domain knowledge, fine-tuning enables them to perform the tasks we anticipate, such as question-answering, dialogue, and reasoning. Given the availability of instruction data, SFT is employed to guide the models in executing semiconductor-related tasks.

\textbf{Post-training process:} Following pre-training and fine-tuning, we conducted quantization and merging to prepare the models for deployment. Our implementation utilized GPTQ [30], an accurate post-training quantization technique for generative pre-trained transformers. Finally, the LoRA adapter was merged with the original LLM model to produce the final LLM model tailored for semiconductor manufacturing.

\begin{table}[h!]
\centering
\caption{Overview of models}
\begin{tabularx}{\textwidth}{>{\centering\arraybackslash}X>{\centering\arraybackslash}X>{\centering\arraybackslash}X}
\toprule
\textbf{Feature} & \textbf{SemiKong-7B} & \textbf{SemiKong-80B} \\ \midrule
Number of trainable parameters & 2,849,712 & 50,732,193 \\ \midrule
Hardware Resources for Training & NVIDIA GPU 4x A100 80GB & NVIDIA GPU 8x A100 80GB \\ \midrule
Training time & \(\sim\)150 hours (15 runs) & \(\sim\)200 hours (2 runs) \\ \bottomrule
\end{tabularx}
\label{table:models}
\end{table}

\subsection{The proposed method for evaluating  LLM in semiconductor manufacturing}
\label{expert_feedback_section}

\begin{figure}[h!]
\centering
\includegraphics[width=0.95\textwidth]{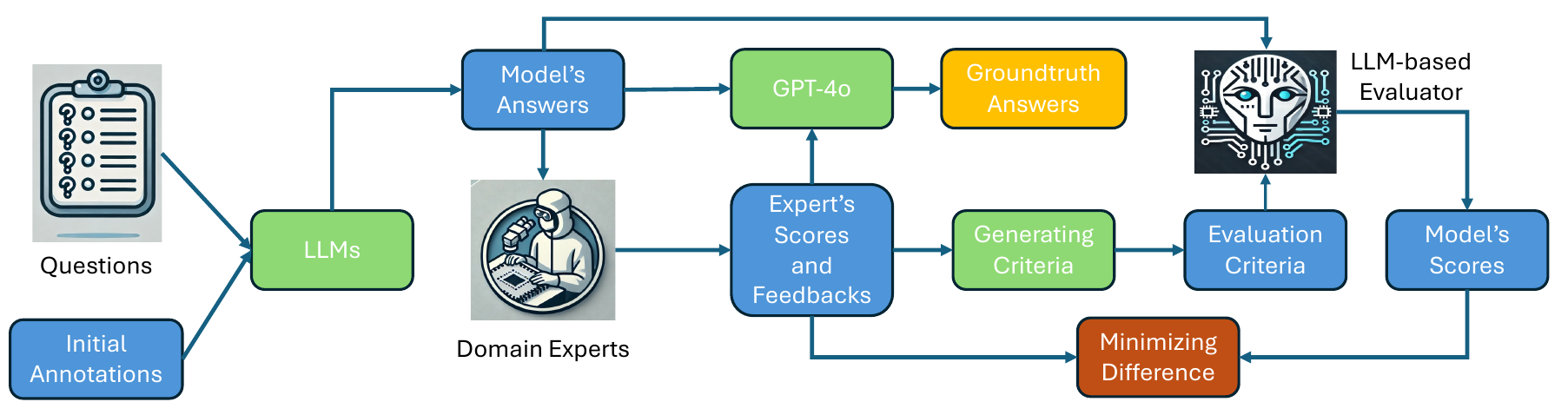} 
\caption{\textbf{The evaluation benchmark development pipeline.}}
\label{fig:evaluation-pipeline}
\end{figure}

The evaluation of AI assistant models in domain-specific contexts requires expert judgment to justify the usefulness of the model's responses. However, expert annotations are often limited and costly. Therefore, developing an automated metric to assess the quality of these models is crucial for their development and evaluation. Such a metric not only supports project development but also serves as a standard for future research in this area. Motivated by this need, we propose a novel pipeline to generate a list of criteria for evaluation. This list of criteria will be fed into LLMs to enhance their ability to justify expert models. A key challenge is that different subfields require different evaluation criteria, and no universal criteria apply to all problems. We anticipate that with the finalized list of criteria, LLMs will be able to evaluate the responses from AI assistant models with a high correlation to expert judgment. Our contribution includes developing a pipeline for generating a customized list of criteria by leveraging expert feedback. We demonstrate the effectiveness of our pipeline by generating a list of criteria for the semiconductor industry domain. It is important to emphasize that our method is not only applicable to the semiconductor domain but also to other domains requiring human expertise.
In our proposed pipeline for evaluation, we initially collected a set of questions from three primary sources: 737 questions from our company’s experts, 150 questions crawled from the ResearchGate forum, and 100 general questions generated by ChatGPT. Our internal experts carefully reviewed and evaluated each question to ensure its quality. Following this review, questions were classified into three difficulty levels: Easy, Medium, and Hard, as detailed in Table~\ref{tab:difficulty_levels_balanced}. Additionally, our experts developed an ontology, as detailed in Section~\ref{ontology_section}, to categorize the questions' processes into high, sub, and specific levels. Finally, we utilized all the collected questions and annotations, inputting them into GPT-4o and our SemiKong model to generate the initial answers.

\begin{table}[h!]
\centering
\caption{Difficulty levels of questions in the evaluation dataset}
\begin{tabularx}{0.8\textwidth}{>{\centering\arraybackslash}X>{\centering\arraybackslash}X>{\centering\arraybackslash}X>{\centering\arraybackslash}X}
\toprule
\textbf{Level of difficulty} & \textbf{Easy} & \textbf{Medium} & \textbf{Hard} \\ \midrule
\textbf{\#Questions} & 100 & 737 & 150 \\ \bottomrule
\end{tabularx}
\label{tab:difficulty_levels_balanced}
\end{table}

Building upon the human-in-the-loop concept, we have advanced it to an expert-in-the-loop framework. As shown in Figure II, in this approach, experts review the initial answers generated by LLMs. These experts, who possess extensive knowledge in their fields, not only provide correct answers but also evaluate the quality of other answers. This dual capability allows us to generate ground truth for benchmarking and to synthesize a set of criteria to guide LLMs in evaluating semiconductor expert models. To implement this, we request experts to score the answers and provide detailed justifications for their scores. Machine learning researchers then analyze these justifications to develop a comprehensive criteria list, which is used to guide LLMs to score model outputs. The goal is to create clear, precise criteria that enable LLMs to make evaluations similar to those of human experts. This process is iterative, with continuous updates to the criteria based on new data annotations from experts, thereby progressively improving the evaluation framework.
In this paper, we define the criteria for using LLMs to evaluate semiconductor expert models as follows:

\textbf{Clarity and Directness (C\&D):} This criterion involves using simple and straightforward language to ensure that the answer is easily understood. This means avoiding unnecessary jargon or technical terms that could confuse the reader. It also requires directly addressing the question or topic at hand in each sentence, maintaining a focus on the main points. Organizing information with bullet points or numbered lists can further enhance readability and make key points more accessible.

\textbf{Practicality and Immediate Usability (PIU):} Practicality and immediate usability involve providing recommendations that are both practical and easy to implement. This means focusing on clear, actionable steps rather than theoretical explanations, ensuring that the guidance is directly applicable to real-world situations. Recommendations should be realistic and suitable to the specific context, making them immediately usable and relevant to the audience's needs.

\textbf{Efficiency and Brevity (E\&B):} Efficiency and brevity involve eliminating redundant information and combining related points to avoid verbosity. The goal is to keep the information concise while still covering all necessary details, ensuring that the message is clear and to the point without unnecessary elaboration.

\textbf{Logical Flow and Coherence (LFC):} Logical flow and coherence involve arranging points in a clear, logical order to make the answer easy to follow. This includes grouping related points together under clear categories, enhancing the overall coherence, and ensuring that the user can easily understand the progression of ideas.

\textbf{Expert-to-Expert Communication (EEC):} Expert-to-expert communication involves tailoring responses as instructions or directions that an experienced engineer would give to another engineer in the same role but with less experience. This ensures that the conversation is part of a problem-solving process, focusing on advanced concepts and practical guidance without delving into overly basic explanations that would be unnecessary for an expert audience.

\textbf{Use of Examples and Specificity (UES):} The use of examples and specificity involves providing examples only when they add significant value to the explanation. Ensure that comparisons are directly related to the point being made and are concise. Introduce technical terms only if they are essential to the discussion, and offer concise explanations for these terms only if requested to maintain clarity and relevance.



\section{Experimental Result}

\label{experiment_section}

\subsection{Implementation details}
For training Semikong, we utilized 8 NVIDIA A100 80GB GPUs. We followed guidelines from Transformers HuggingFace, HuggingFace Accelerator, and the LLaMA-Factory library for fine-tuning a LLM. The hyperparameters for pretraining and SFT included a batch size of 3, gradient accumulation steps of 3, and a learning rate of 1.0e-5. The training was conducted over 5 epochs, employing a cosine learning rate scheduler with a warm-up ratio of 0.15. We enabled FP16 for mixed-precision training and allocated 20\% of the dataset for validation. LoRA was employed in our finetuning.

\subsection{Evaluation}

To evaluate the contribution of finetuning and pre-training, we conducted experiments to compare three models: Llama3, SemiKong SFT only, and pretrained SemiKong with SFT. Table~\ref{tab:sft_contribution} shows the result of our experiment. In general, fine tuning only did not improve the performance of the model. It shows that generic models lack knowledge of domain specifics. When the model is pre-trained to learn more in-depth knowledge, the model's performance begins to show signs of improvement. However, the model implemented for this experiment only has 8B parameters, which limits the ability to learn knowledge of the model. So, in the next experiments, we will conduct the experiment on a larger model with 70B parameters and fine-tune only the model that has been pre-trained with our proposed semiconductor dataset.

\begin{table*}[h!]
\centering
\caption{Comparison of contribution of SFT and pre-training}
\begin{tabularx}{\textwidth}{l *{7}{>{\centering\arraybackslash}X}}
\toprule
\textbf{Model} & \textbf{C\&D} & \textbf{PIU} & \textbf{E\&B} & \textbf{LFC} & \textbf{EEC} & \textbf{UES} & \textbf{Total} \\ \midrule
Llama3 8B & 3.65 & 3.35 & 3.07 & 3.67 & 3.47 & \textbf{3.28} & 20.49 \\ 
SemiKong 8B (SFT only) & 3.61 & \textbf{3.36} & 3.22 & 3.64 & 3.52 & 3.16 & 20.51 \\ 
SemiKong 8B (Pretraining+SFT) & \textbf{3.73} & 3.35 & \textbf{3.40} & \textbf{3.68} & \textbf{3.54} & 3.11 & \textbf{20.81} \\ 
\bottomrule
\end{tabularx}
\label{tab:sft_contribution}
\end{table*}

\begin{table*}[h!]
\centering
\caption{Compare with open source models}
\begin{tabularx}{\textwidth}{l *{7}{>{\centering\arraybackslash}X}}
\toprule
\textbf{Model} & \textbf{C\&D} & \textbf{PIU} & \textbf{E\&B} & \textbf{LFC} & \textbf{EEC} & \textbf{UES} & \textbf{Total} \\ \midrule
Llama3 8B & 3.65 & 3.35 & 3.07 & 3.67 & 3.47 & 3.28 & 20.49 \\ 
SemiKong 8B & 3.73 & 3.35 & 3.40 & 3.68 & 3.54 & 3.11 & 20.81 \\ 
Llama3 70B & 3.89 & 3.63 & 3.55 & 3.99 & 3.82 & 3.47 & 22.35 \\ 
SemiKong 70B (Pretraining+SFT) & \textbf{4.07} & \textbf{4.05} & \textbf{3.88} & \textbf{4.23} & \textbf{4.13} & \textbf{3.66} & \textbf{24.02} \\ 
\bottomrule
\end{tabularx}
\label{tab:open_source_models}
\end{table*}

The experimental results in Table~\ref{tab:open_source_models} demonstrate that models with 70B parameters significantly outperform those with 8B parameters. Even when compared to our fine-tuned SemiKong 8B model, the base Llama3 70B model still outperforms it. Based on this observation, our SemiKong 70B model and the experimental results show that our approach significantly surpasses both the generic open-source Llama3 8B and Llama3 70B models across all criteria.

\begin{table*}[h!]
\centering
\caption{Compare with commercial products}
\begin{tabularx}{\textwidth}{l *{7}{>{\centering\arraybackslash}X}}
\toprule
\textbf{Model} & \textbf{C\&D} & \textbf{PIU} & \textbf{E\&B} & \textbf{LFC} & \textbf{EEC} & \textbf{UES} & \textbf{Total} \\ \midrule
Claude-3.5-Sonnet & 3.80 & 3.44 & 3.15 & 3.82 & 3.67 & 3.37 & 21.25 \\ 
Claude-3-haiku & \textbf{3.95} & 3.54 & \textbf{3.64} & 3.92 & 3.80 & 3.31 & 22.16 \\ 
Claude-3-opus & 3.88 & 3.47 & 3.50 & 3.86 & 3.75 & 3.29 & 21.75 \\ 
Command-R & 3.76 & 3.46 & 3.38 & 3.74 & 3.64 & 3.22 & 21.20 \\ 
GPT-3.5 & 3.32 & 2.86 & 3.05 & 3.08 & 3.00 & 2.59 & 17.90 \\ 
SemiKong-70B & 3.87 & \textbf{3.84} & 3.59 & \textbf{3.99} & \textbf{3.90} & \textbf{3.46} & \textbf{22.65} \\ 
\bottomrule
\end{tabularx}
\label{tab:commercial_products_comparison}
\end{table*}

To demonstrate the superiority of SemiKong, we conducted experiments comparing its performance with that of commercial products. It's important to note that SemiKong is a foundation model and does not rely on supporting systems like RAG. As shown in Table~\ref{tab:commercial_products_comparison} and Figure~\ref{fig:products_comparison}, SemiKong delivers comparable performance in the C\&D and E\&B metrics, while it outperforms in four out of six key metrics: PIU, LFC, EEC, and UES. These metrics are critical for determining whether a model meets the needs of an expert. Overall, SemiKong achieves state-of-the-art performance, making it the most suitable model for expert use. Its practicality for immediate application, logical flow, avoidance of unnecessary information, and ability to provide concise and accurate answers are exactly what engineers require for their daily work.

\section{Conclusion and Future Research Directions}
\label{discussion_section}

In this paper, we introduce SemiKong, the first foundation model specialized for the semiconductor industry, available in both 8B and 70B versions. Additionally, we have publicized a large-scale dataset tailored for semiconductor applications, encompassing both pretraining and fine-tuning data. We also present a semiconductor ontology designed to support AI researchers in developing new AI research within the semiconductor field. Our SemiKong models have achieved state-of-the-art performance, outperforming open-source foundation models and surpassing commercial products in expert use. However, SemiKong represents just the initial effort, and there remains significant work to be done. First, based on our proposed ontology, we can further develop additional processes beyond etching, making AI for semiconductors more comprehensive and applicable to various stages of semiconductor manufacturing. Secondly, our pipeline can be adapted and expanded to other industries, thereby enhancing industrial operations across multiple sectors.

\section*{Acknowledgments}

We would like to express our gratitude to the AI Alliance (\url{https://thealliance.ai}) for providing the impetus, resources, and platform for this work, and for collaboration in open science. We also extend our thanks to the member organizations of the AI Alliance, their researchers, and engineers for their valuable contributions to this study, including Anthony Annunziata (IBM Research), Sean Hughes (ServiceNow), Phong Nguyen (FPT Software, AI Center), Noritaka Yokomori (Tokyo Electron). Their expertise, insights, and collaborative spirit have been instrumental in advancing our research.

\bibliographystyle{unsrtnat}
\bibliography{references}

\appendix
\section*{Appendix}

\section{Semiconductor Process Ontology}
\label{ontology_section_appendix}

\begin{longtable}{|p{5cm}|p{4cm}|p{6cm}|}
\caption{Detailed Semiconductor Process Overview} \label{table:semiconductor_process} \\

\hline
\rowcolor{cyan!20}
\textbf{Process Group} & \textbf{Process Module} & \textbf{Process Unit} \\ 
\hline
\endfirsthead

\hline
\rowcolor{cyan!20}
\textbf{Process Group} & \textbf{Process Module} & \textbf{Process Unit} \\ 
\hline
\endhead

\hline
\multicolumn{3}{|r|}{\textit{Continued on next page}} \\ 
\hline
\endfoot

\hline
\endlastfoot

\multirow{3}{*}{1. Substrate Preparation} & 1.1 Wafer Manufacturing & 1.1.1 Crystal Growth \\ \cline{3-3}
 & & 1.1.2 Wafer Slicing \\ \cline{3-3}
 & & 1.1.3 Edge Rounding \\ \cline{2-3}
 & 1.2 Wafer Polishing & 1.2.1 Lapping \\ \cline{3-3}
 & & 1.2.2 Chemical Mechanical Polishing \\ \cline{2-3}
 & 1.3 Cleaning & 1.3.1 RCA Clean \\ \cline{3-3}
 & & 1.3.2 Piranha Clean \\ \cline{3-3}
 & & 1.3.3 Vapor Phase Cleaning \\ \hline
\multirow{2}{*}{2. Film Formation} 
& 2.1 Oxidation 
& 2.1.1 Thermal Oxidation \\ \cline{3-3}
& & 2.1.2 Plasma-Enhanced Oxidation \\ \cline{3-3}
& & 2.1.3 High Pressure Oxidation \\ \cline{3-3}
& & 2.1.4 Low Pressure Oxidation \\ \cline{3-3}
& & 2.1.5 Anodic Oxidation \\ \cline{2-3}
& 2.2 Deposition 
& 2.2.1 Chemical Vapor Deposition (CVD) \\ \cline{3-3}
& & 2.2.2 Physical Vapor Deposition (PVD) \\ \cline{3-3}
& & 2.2.3 Atomic Layer Deposition (ALD) \\ \cline{3-3}
& & 2.2.4 Pulsed Layer Deposition \\ \cline{2-3}
& 2.3 Epitaxial Growth 
& 2.3.1 Silicon Epitaxy \\ \cline{3-3}
& & 2.3.2 Compound Semiconductor Epitaxy \\ \hline
\multirow{2}{*}{3. Patterning} 
& 3.1 Lithography 
& 3.1.1 Photoresist Application \\ \cline{3-3}
& & 3.1.2 Exposure (UV Deep UV EUV) \\ \cline{3-3}
& & 3.1.3 Development \\ \cline{3-3}
& & 3.1.4 Electron Beam Lithography \\ \cline{3-3}
& & 3.1.5 Ion Beam Lithography \\ \cline{3-3}
& & 3.1.6 Maskless Lithography \\ \cline{3-3}
& & 3.1.7 Immersion Lithography \\ \cline{2-3}
& 3.2 Etching 
& 3.2.1 Wet Etching \\ \cline{3-3}
& & 3.2.2 Dry Etching \\ \cline{3-3}
& & 3.2.3 Plasma Etching \\ \cline{3-3}
& & 3.2.4 Reactive Ion Etching \\ \cline{3-3}
& & 3.2.5 Deep Reactive Ion Etching \\ \cline{3-3}
& & 3.2.6 Isotropic Wet Etching \\ \cline{3-3}
& & 3.2.7 Anisotropic Wet Etching \\ \cline{3-3}
& & 3.2.8 Atomic Ion Etching \\ \cline{3-3}
& & 3.2.9 Electron Cyclotron Etching \\ \hline
\multirow{2}{*}{4. Doping} 
& 4.1 Ion Implantation 
& 4.1.1 High Energy Implantation \\ \cline{3-3}
& & 4.1.2 Low Energy Implantation \\ \cline{3-3}
& & 4.1.3 Plasma Immersion Ion Implantation \\ \cline{3-3}
& & 4.1.4 Focused Ion Beam Implantation \\ \cline{2-3}
& 4.2 Diffusion 
& 4.2.1 Thermal Diffusion \\ \cline{3-3}
& & 4.2.2 Rapid Thermal Diffusion \\ \cline{2-3}
& 4.3 In-situ Doping 
& 4.3.1 During Epitaxial Growth \\ \cline{3-3}
& & 4.3.2 During Deposition \\ \hline
\multirow{2}{*}{5. Planarization} 
& 5.1 Chemical Mechanical Planarization 
& 5.1.1 Oxide CMP \\ \cline{3-3}
& & 5.1.2 Metal CMP \\ \cline{2-3}
& 5.2 Etchback Planarization 
& 5.2.1 Resist Etchback \\ \cline{3-3}
& & 5.2.2 Sacrificial Layer Etchback \\ \hline

\multirow{3}{*}{6. Cleaning and Surface Preparation} 
& 6.1 Wet Cleaning 
& 6.1.1 RCA Clean \\ \cline{3-3}
& & 6.1.2 Piranha Clean \\ \cline{3-3}
& & 6.1.3 HF Dip \\ \cline{2-3}
& 6.2 Dry Cleaning 
& 6.2.1 Plasma Cleaning \\ \cline{3-3}
& & 6.2.2 UV-Ozone Cleaning \\ \cline{2-3}
& 6.3 Advanced Cleaning 
& 6.3.1 Supercritical CO2 Cleaning \\ \cline{3-3}
& & 6.3.2 Cryogenic Cleaning \\ \hline

\multirow{3}{*}{7. Thermal Processing} 
& 7.1 Annealing 
& 7.1.1 Furnace Annealing \\ \cline{3-3}
& & 7.1.2 Rapid Thermal Annealing \\ \cline{3-3}
& & 7.1.3 Laser Annealing \\ \cline{2-3}
& 7.2 Thermal Oxidation 
& 7.2.1 Dry Oxidation \\ \cline{3-3}
& & 7.2.2 Wet Oxidation \\ \cline{2-3}
& 7.3 Dopant Activation 
& 7.3.1 Spike Annealing \\ \cline{3-3}
& & 7.3.2 Flash Annealing \\ \hline

\multirow{3}{*}{8. Metrology and Inspection} 
& 8.1 Physical Metrology 
& 8.1.1 Profilometry \\ \cline{3-3}
& & 8.1.2 Ellipsometry \\ \cline{3-3}
& & 8.1.3 X-ray Reflectometry \\ \cline{2-3}
& 8.2 Electrical Metrology 
& 8.2.1 Sheet Resistance Measurement \\ \cline{3-3}
& & 8.2.2 Capacitance-Voltage Measurement \\ \cline{2-3}
& 8.3 Defect Inspection 
& 8.3.1 Optical Inspection \\ \cline{3-3}
& & 8.3.2 E-beam Inspection \\ \cline{3-3}
& & 8.3.3 Wafer Inspection \\ \hline
\multirow{3}{*}{9. Advanced Modules} 
& 9.1 High-k/Metal Gate 
& 9.1.1 Gate Dielectric Deposition \\ \cline{3-3}
& & 9.1.2 Metal Gate Deposition \\ \cline{2-3}
& 9.2 Strain Engineering 
& 9.2.1 Strained Silicon \\ \cline{3-3}
& & 9.2.2 SiGe Channels \\ \cline{2-3}
& 9.3 3D Structures 
& 9.3.1 FinFET Formation \\ \cline{3-3}
& & 9.3.2 Gate-All-Around Structures \\ \hline
\multirow{9}{*}{10. Back-End Processes} 
& 10.1 Multilayer Interconnect 
& 10.1.1 Interlayer Dielectric Deposition \\ \cline{3-3}
& & 10.1.2 Metal Deposition \\ \cline{3-3}
& & 10.1.3 Chemical Mechanical Planarization \\ \cline{2-3}
& 10.2 Metallization 
& 10.2.1 Physical Vapor Deposition \\ \cline{3-3}
& & 10.2.2 Chemical Vapor Deposition \\ \cline{3-3}
& & 10.2.3 Electroplating \\ \cline{3-3}
& & 10.2.4 Sputtering \\ \cline{2-3}
& 10.3 Interconnect Patterning 
& 10.3.1 Damascene Process \\ \cline{3-3}
& & 10.3.2 Dual Damascene Process \\ \cline{2-3}
& 10.4 Passivation 
& 10.4.1 Silicon Nitride Deposition \\ \cline{3-3}
& & 10.4.2 Polyimide Coating \\ \cline{2-3}
& 10.5 Wafer Thinning 
& 10.5.1 Backside Grinding \\ \cline{3-3}
& & 10.5.2 Chemical Etching \\ \cline{2-3}
& 10.6 Wafer Testing 
& 10.6.1 Parametric Testing \\ \cline{3-3}
& & 10.6.2 Functional Testing \\ \cline{2-3}
& 10.7 Dicing 
& 10.7.1 Mechanical Dicing \\ \cline{3-3}
& & 10.7.2 Laser Dicing \\ \cline{3-3}
& & 10.7.3 Plasma Dicing \\ \cline{2-3}
& 10.8 Packaging 
& 10.8.1 Die Attach \\ \cline{3-3}
& & 10.8.2 Wire Bonding \\ \cline{3-3}
& & 10.8.3 Flip Chip Bonding \\ \cline{3-3}
& & 10.8.4 Encapsulation \\ \cline{2-3}
& 10.9 Advanced Packaging 
& 10.9.1 Through-Silicon Via (TSV) \\ \cline{3-3}
& & 10.9.2 Wafer-Level Packaging \\ \cline{3-3}
& & 10.9.3 3D Integration \\ \hline

\end{longtable}

\end{document}